\documentclass{article}

\usepackage{microtype}
\usepackage{graphicx}
\usepackage{booktabs} % for professional tables

\usepackage{hyperref}

\usepackage[accepted]{icml2024}

\usepackage{amsmath}
\usepackage{amssymb}
\usepackage{mathtools}
\usepackage{amsthm}
\usepackage{subcaption}
\usepackage{multirow}

\usepackage[capitalize,noabbrev]{cleveref}
\usepackage[acronym]{glossaries}
\glsdisablehyper
\newacronym{ame}{AME}{attention and memory enhancement}
\newacronym{bert}{BERT}{bidirectional encoder representations from transformers}
\newacronym{boot}{BooT}{bootstrapped transformer}
\newacronym{cnn}{CNN}{convolutional neural network}
\newacronym{cv}{CV}{computer vision}
\newacronym{dag}{DAG}{directed acyclic graph}
\newacronym{dgerd}{DGERD}{disjunctive graph embedded recurrent decoding}
\newacronym{dt}{DT}{decision transformer}
\newacronym{esper}{ESPER}{environment-stochasticity-independent representations}
\newacronym{gan}{GAN}{generative adversarial network}
\newacronym{gnn}{GNN}{graph neural network}
\newacronym{gpt}{GPT}{generative pre-trained transformer}
\newacronym{gru}{GRU}{gated recurrent unit}
\newacronym{gtr-xl}{GTrXL}{gated transformer-XL}
\newacronym{happo}{HAPPO}{heterogeneous-agent proximal policy optimization}
\newacronym{hpo}{HPO}{hyper-parameter optimization}
\newacronym{iid}{i.i.d}{independent and identically distributed}
\newacronym{iot}{IoT}{internet of things}
\newacronym{iris}{IRIS}{imagination with auto-regression over an inner speech}
\newacronym{llm}{LLM}{large language model}
\newacronym{lstm}{LSTM}{long short-term memory}
\newacronym{maans}{MAANS}{multi-agent active neural SLAM}
\newacronym{maml}{MAML}{model-agnostic meta-learning}
\newacronym{mappo}{MAPPO}{multi-agent proximal policy optimization}
\newacronym{marl}{MARL}{multi-agent reinforcement learning}
\newacronym{mbrl}{MBRL}{model-based reinforcement learning}
\newacronym[longplural=Markov decision processes]{mdp}{MDP}{Markov decision process}
\newacronym{mlp}{MLP}{multi-layer perceptron}
\newacronym{mse}{MSE}{mean-squared error}
\newacronym{mtrl}{MTRL}{multi-task reinforcement learning}
\newacronym{nlp}{NLP}{natural language processing}
\newacronym{pomdp}{POMDP}{partially observable Markov decision process}
\newacronym{mwm}{MWM}{Masked world model}
\newacronym{ppo}{PPO}{proximal policy optimization}
\newacronym{qdt}{QDT}{$Q$-learning decision transformer}
\newacronym{rat}{RAT}{relation-aware transformer}
\newacronym{rl}{RL}{reinforcement learning}
\newacronym{rlhf}{RLHF}{reinforcement learning with human feedback}
\newacronym{rnn}{RNN}{recurrent neural network}
\newacronym{td}{TD}{temporal difference}
\newacronym{tr-xl}{TrXL}{transformer-XL}
\newacronym{tt}{TT}{trajectory transformer}
\newacronym{vit}{ViT}{vision transformer}
\newacronym{dtd}{DTd}{decision transducer}
\newacronym{rssm}{RSSM}{recurrent state space model}
\newacronym{twm}{TWM}{transformer-based world model}
\newacronym{dtr}{DTR}{discrete-tokenized representations}
\newacronym{vq-vae}{VQ-VAE}{vector quantized-variational autoencoder}
\newacronym{dart}{DART}{discrete abstract representation for transformer-based learning}
\newacronym{psnr}{PSNR}{peak signal-to-noise ratio}
\newacronym{mcts}{MCTS}{Monte Carlo tree search}

\theoremstyle{plain}

\theoremstyle{definition}

\theoremstyle{remark}

\usepackage[textsize=tiny]{todonotes}
\graphicspath{ {./figs/} }

\begin{document}

\twocolumn[
\icmltitle{Learning to Play Atari in a World of Tokens}

\icmlsetsymbol{equal}{*}

\begin{icmlauthorlist}
\icmlauthor{Pranav Agarwal}{comp,yyy}
\icmlauthor{Sheldon Andrews}{comp,rbx}
\icmlauthor{Samira Ebrahimi Kahou}{uoc,yyy,sch}
\end{icmlauthorlist}

\icmlaffiliation{comp}{École de Technologie Supérieure, Canada}
\icmlaffiliation{uoc}{University of Calgary, Canada}
\icmlaffiliation{yyy}{Mila}
\icmlaffiliation{sch}{Canada CIFAR AI Chair}
\icmlaffiliation{rbx}{Roblox, USA}

\icmlcorrespondingauthor{Pranav Agarwal}{pranav.agarwal.1@ens.etsmtl.ca}

\icmlkeywords{Machine Learning, ICML}

\vskip 0.3in
]

\printAffiliationsAndNotice{}  

\begin{abstract}
Model-based reinforcement learning agents utilizing transformers have shown improved sample efficiency due to their ability to model extended context, resulting in more accurate world models.
However, for complex reasoning and planning tasks, these methods primarily rely on continuous representations.
This complicates modeling of discrete properties of the real world such as disjoint object classes between which interpolation is not plausible.
In this work, we introduce discrete abstract representations for transformer-based learning (DART), a sample-efficient method utilizing discrete representations for modeling both the world and learning behavior. We incorporate a transformer-decoder for auto-regressive world modeling and a transformer-encoder for learning behavior by attending to task-relevant cues in the discrete representation of the world model. For handling partial observability, we aggregate information from past time steps as memory tokens.  
DART outperforms previous state-of-the-art methods that do not use look-ahead search on the Atari 100k sample efficiency benchmark with a median human-normalized score of 0.790 and beats humans in 9 out of 26 games. We release our code at \href{https://pranaval.github.io/DART/}{https://pranaval.github.io/DART/}.
\end{abstract}

\section{Introduction}
\label{submission}

A \gls{rl} algorithm usually takes millions of trajectories to master a task, and the training can take days or even weeks, especially when using complex simulators. This is where \gls{mbrl} comes in handy~\citep{sutton1991dyna}. With \gls{mbrl}, the agent learns the \textit{dynamics} of the environment, understanding how the environment state changes when different actions are taken~\citep{10007800}. This method is more efficient because the agent can train in its \textit{imagination} without requiring direct interaction with an external simulator or the real environment.~\citep{ha2018world}. Additionally, the learned model allows the agent for safe and accurate decision-making by utilizing different look-ahead search algorithms for planning its action~\citep{hamrick2020role}.

Most \gls{mbrl} methods commonly follow a structured three-step approach: 1) Representation Learning \( \phi : S \rightarrow \mathbb{R}^n \), the agents capture a simplified representation $\mathbb{R}^n$ of the high dimensional environment state $S$; 2) Dynamics and Reward Learning \( f : S \times A \rightarrow S', \psi : S \times A \times S' \rightarrow R \), where the agent learns the dynamics of the environment, predicting the next state \( s' \) given the current state \( s \) and action \( a \), as well as the reward associated with transitioning from \( s \) to \( s' \); and 3) Policy Learning \( \pi : S \rightarrow \mathcal{P}(A) \), the agent determines the optimal actions needed to achieve its goals. Dreamer is a family of \gls{mbrl} agents that follow a similar structured three-step approach.

DreamerV1~\citep{hafner2019dream} employed a \gls{rssm}~\citep{doerr2018probabilistic} to learn the world model. DreamerV2~\citep{hafner2020mastering}, an improved version of DreamerV1, offers better sample efficiency and scalability by incorporating a discrete latent space for modeling the dynamics. Building on the advancements of DreamerV2, DreamerV3~\citep{hafner2023mastering} takes a similar approach with additions involving the use of symlog predictions and various regularisation techniques aimed at stabilizing learning across diverse environments. Notably, DreamerV3 surpasses the performance of past models across a wide range of tasks, while using fixed hyperparameters.

Although Dreamer variants are among the most popular \gls{mbrl} approaches, they suffer from sample-inefficiency~\citep{yin2022planning, svidchenko2021maximum}. The training of Dreamer models can require an impractical amount of gameplay time, ranging from months to thousands of years, depending on the complexity of the game~\citep{micheli2022transformers}. This inefficiency can be primarily attributed to inaccuracies in the learned world model, which tend to propagate errors into the policy learning process, resulting in compounding error problems~\citep{xiao2019learning}. This challenge is largely associated with the use of \glspl{cnn} and \glspl{rnn}~\citep{deng2023facing} that, while effective in many domains, face limitations in capturing complex and long-range dependencies, which are common in \gls{rl} scenarios~\citep{ni2024transformers}.

\begin{figure*}[!t]
    \centering
    \includegraphics[width=0.95\linewidth]{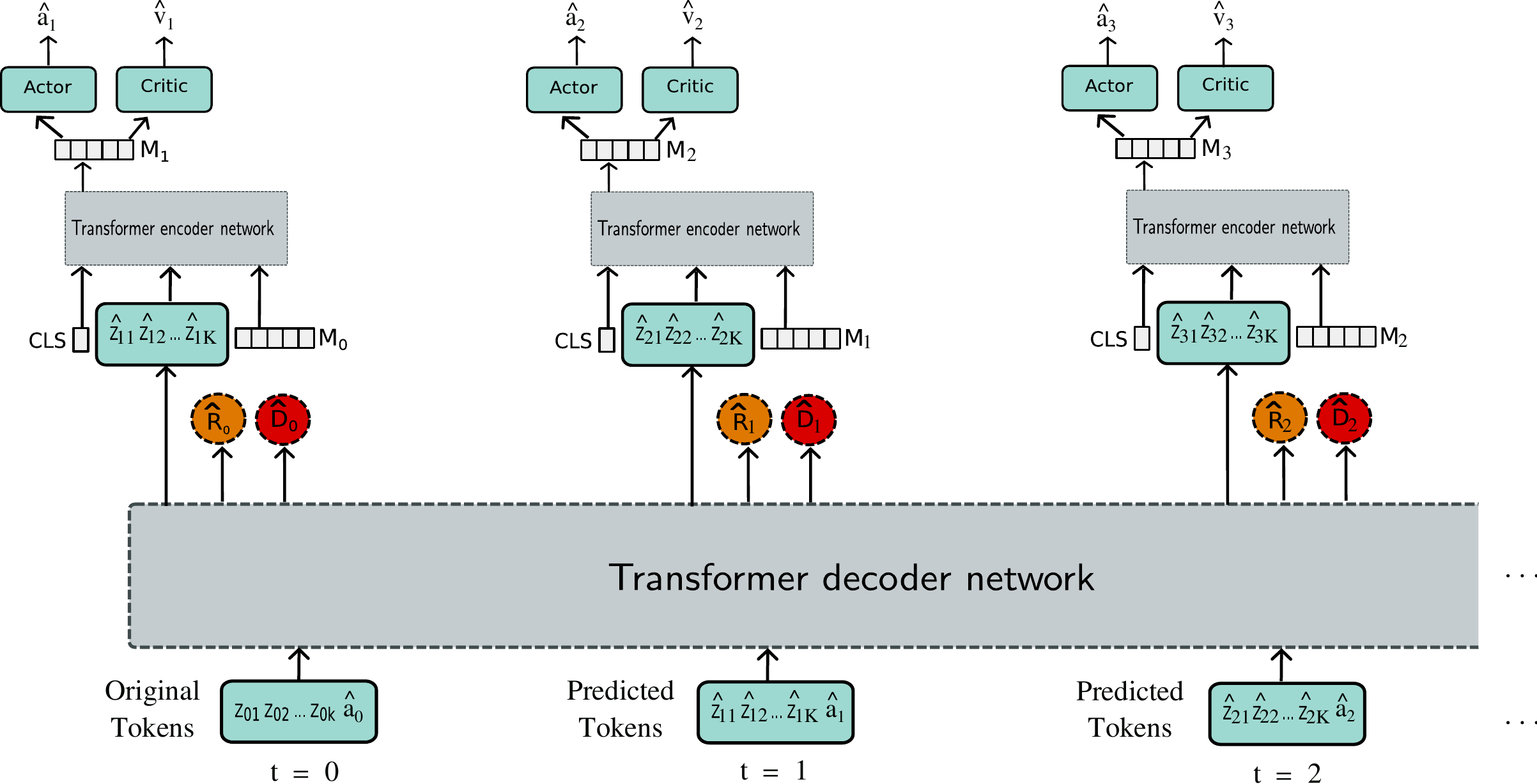}
    \caption{Discrete abstract representation for transformer-based learning (DART): In this approach, the original observation $x_t$ is encoded into discrete tokens $z_t$ using VQ-VAE. These tokenized observations, and predicted action, serve as inputs for the world model. A Transformer decoder network is used for modeling the world. The predicted tokens, along with a \texttt{CLS} and a \texttt{MEM} token are used as input by the policy. This policy is modeled using a transformer-encoder network. The \texttt{CLS} token aggregates information from the observation tokens and the \texttt{MEM} token to learn a common representation, which is then used for action and value predictions. This common representation also plays a role in modeling memory, acting as the \texttt{MEM} token at the subsequent time step.}
    \label{fig:my_figure}
\end{figure*}

This motivates the need to use transformers~\citep{vaswani2017attention, lin2022survey}, which have proven highly effective in capturing long-range dependencies in various \gls{nlp} tasks~\citep{wolf2020transformers} and addressing complex visual reasoning challenges in \gls{cv} tasks~\citep{khan2022transformers}. Considering these advantages, recent works have adapted transformers for modeling the dynamics in \gls{mbrl}. Transdreamer~\citep{chen2022transdreamer} first used a transformer-based world model by replacing Dreamer's \gls{rnn}-based stochastic world model with a transformer-based state space model. It outperformed DreamerV2 in Hidden Order Discovery Tasks which requires long-term dependency and complex-reasoning. In order to stabilize the training, it utilizes \gls{gtr-xl}~\citep{parisotto2020stabilizing} architecture.

\gls{mwm}~\citep{seo2023masked} utilizes a convolutional-autoencoder and \gls{vit}~\citep{dosovitskiy2020image} for learning a representation that models dynamics following the \gls{rssm} objective. Their decoupling approach outperforms DreamerV2 on different robotic manipulation tasks from Meta-world~\citep{yu2020meta} and RLBench~\citep{james2020rlbench}. Similarly, \gls{twm}~\citep{robine2023transformer} use \gls{tr-xl}~\citep{dai2019transformer} for modeling the world and use the predicted latent states for policy learning. Their work demonstrates sample-efficient performance on the Atari 100k benchmark.

Contrary to these approaches, \gls{iris}~\citep{micheli2022transformers} models dynamics learning as a sequence modeling problem, utilizing discrete image tokens for modeling the world. It then uses reconstructed images using the predicted tokens for learning the policy using \glspl{cnn} and \glspl{lstm}, achieving improved sample efficiency on the Atari 100k compared to past models. However, it still faces difficulties in policy learning due to the use of reconstructed images as input, resulting in reduced performance.

In this work, we introduce \gls{dart}, a novel approach that leverages transformers for learning both the world model and policy. Unlike the previous work by \citet{yoon2023investigation}, which solely utilized a transformer for extracting object-centric representation, our approach employs a transformer encoder to learn behavior through discrete representation ~\citep{mao2021discrete}, as predicted by the transformer-decoder that models the world. This choice allows the model to focus on fine-grained details, facilitating precise decision-making. Specifically, we utilize a transformer-decoder architecture, akin to the \gls{gpt} framework~\citep{radford2019language}, to model the world, while adopting a transformer encoder, similar to the \gls{vit} architecture~\citep{dosovitskiy2020image}, to learn the policy (as illustrated in Figure~\ref{fig:my_figure}).   

Additionally, challenges related to partial observability necessitate memory modeling. Previous work~\cite{didolkar2022temporal} modeled memory in transformers using a computationally intensive two-stream network. Inspired by~\citep{bulatov2022recurrent}, we model memory as a distinct token, aggregating task-relevant information over time using a self-attention mechanism. 

 The main contribution of our work includes a novel approach that utilizes transformers for both world and policy modeling. Specifically, we utilize a transformer-decoder (\gls{gpt}) for world modeling and a transformer-encoder (\gls{vit}) for policy learning. This represents an improvement compared to \gls{iris}, which relies on \glspl{cnn} and \glspl{lstm} for policy learning, potentially limiting its performance. We use discrete representations for policy and world modeling. These discrete representations capture abstract features, enabling our transformer-based model to focus on task-specific fine-grained details. Attending to these details improves decision-making, as demonstrated by our results. To address the problem of partial observability, we introduce a novel mechanism for modeling the memory that aggregates task-relevant information from the previous time step to the next using a self-attention mechanism. Our model showcases enhanced interpretability and sample efficiency. It achieves state-of-the-art results (no-look-ahead search methods) on the Atari 100k benchmark with a median score of 0.790 and superhuman performance in 9 out of 26 games. 
    
\section{Method}

Our model, \gls{dart}, is designed for mastering Atari games, within the framework of a \gls{pomdp}~\citep{kaelbling1998planning} which is defined as a tuple $(\mathcal{O, A}, p, r, \gamma, d)$. Here, $\mathcal{O}$ is the observation space with image observations $x_t \subseteq \mathbb{R}^{h \times w \times 3}$, $\mathcal{A}$ represents the action space, and $a_t$ is a discrete action taken at time step $t$ from the action space $\mathcal{A}$, $p\left(x_t \mid x_{<t}, a_{<t}\right)$ is the transition dynamics, r is the reward function $r_t=r\left(x_{\leq t}, a_{<t}\right)$, $\gamma \in[0,1)$ is the discount factor and $d \in \{0,1\}$ indicates episode termination. The goal is to find a policy $\pi$ that maximizes the expected sum of discounted rewards $\mathbb{E}_\pi\left[\sum_{t=1}^{\infty} \gamma^{t-1} r_t\right]$. Adopting the training methodology employed by \gls{iris}, \gls{dart} likewise consists of three main steps: (1) \textit{Representation Learning}, where \glspl{vq-vae}~\citep{van2017neural,esser2021taming} are used for tokenizing the original observations; (2) \textit{World-Model Learning}, which involves auto-regressive modeling of the dynamics of the environment using a \gls{gpt} architecture; and (3) \textit{Policy Learning}, which is modeled using \gls{vit} for decision-making by attending to task-relevant cues. We now describe our overall approach in detail.

\subsection{Representation Learning}
\label{sec2.1}

Discrete symbols are essential in human communication, as seen in natural languages~\citep{cartuyvels2021discrete}. Likewise, in the context of \gls{rl}, discrete representation is useful for abstraction and reasoning, leveraging the inherent structure of human communication~\citep{islam2022discrete}. This motivates our approach to model the observation space as a discrete set. In this work, we use \gls{vq-vae} for discretizing the observation space. It learns a discrete latent representation of the input data by quantizing the continuous latent space into a finite number of discrete codes,

\begin{equation}
\begin{aligned}
\hat{z}_q = q(\hat{z}_t^k; \phi_q, Z).
\end{aligned}
\label{eq:1}
\end{equation}

At time step $t$, the observation from the environment $x_t \in \mathbb{R}^{H \times W \times 3}$ is encoded by the image encoder $f_{\theta}$ to a continuous latent space $\hat{z}_t^k$. This encoder is modeled using \glspl{cnn}. The quantization process $q$ maps the predicted continuous latent space $\hat{z}_t^k$ to a discrete latent space $\hat{z}_q$. This is done by finding the closest embedding vector in the codebook $Z$ from a set of $N$ codes (see Equation~\ref{eq:1}). The discrete latent codes are passed to the decoder $g_{\phi}$, which maps it back to the input data $\hat{x}_t$.

The training of this \gls{vq-vae} comprises minimizing the \textit{reconstruction loss} to ensure alignment between input and reconstructed images. Simultaneously, the codebook is learned by minimizing the \textit{codebook loss}, encouraging the embedding vector in the codebook to be close to the encoder output. The \textit{commitment loss} encourages the encoder output to be close to the nearest codebook vector. Additionally \textit{perceptual loss} is computed to encourage the encoder to capture high-level features. The total loss in \gls{vq-vae} is a weighted sum of these loss functions.

This approach enables the modeling of fine-grained, low-level information within the input image as a set of discrete latent codes.

\subsection{World-model learning}
The discrete latent representation forms the core of our approach, enabling the learning of dynamics through an autoregressive next-token prediction approach~\citep{qi2024next}. A transformer decoder based on the \gls{gpt} architecture is used for modeling this sequence prediction framework. First, an aggregate sequence $\hat{z}_{ct} = f_{\phi}(\hat{z}_{<t}, \hat{a}_{<t})$ is modeled by encoding past latent tokens and actions at each time step. 
The aggregated sequence is used for estimating the distribution of the next token, contributing to the modeling of future states given as $\hat{z}_{q_{t}}^k \sim p_{d}(\hat{z}_{q_{t}}^k \mid \hat{z}_{ct})$. Simultaneously, it is also used for estimating the reward $\hat{r}_{t} \sim p_{d}(\hat{r}_t \mid \hat{z}_{ct})$ and the episode termination $\hat{d}_{t} \sim p_{d}(\hat{d}_t \mid \hat{z}_{ct})$. This training occurs in a self-supervised manner, with the next state predictor and termination modules trained using cross-entropy loss, while reward prediction uses mean squared error.

\subsection{Policy-learning}
The policy $\pi$ is trained within the world model (also referred as imagination) using a transformer encoder architecture based on \glsfirst{vit}. At each time step $t$, the policy processes the current observation as $K$ discrete tokens received from the world model. These observation tokens are extended with additional learnable embeddings, including a \texttt{CLS} token placed at the beginning and a \texttt{MEM} token appended to the end, 

\begin{equation}
\begin{aligned}
& \mathbf{out} = [\texttt{CLS}, \hat{z}_{q_{t}}^{1}, \ldots, \hat{z}_{q_{t}}^{K}, \texttt{MEM}_{t-1}] + \mathbf{E}_{\text{pos}}. \\
\end{aligned}
\label{eq:lab9}
\end{equation}

The \texttt{CLS} token helps in aggregating information from the $K$ observation tokens and the \texttt{MEM} token. Meanwhile, the \texttt{MEM} token acts as a memory unit, accumulating information from the previous time steps. Thus, at time step $t$ the input to the policy can be represented as ($\texttt{CLS}, \hat{z}_{q_{t}}^{1}, \ldots, \hat{z}_{q_{t}}^{K}, \texttt{MEM}_{t-1}$), where $\hat{z}_{q_{t}}^{K}$ corresponds to the embedding of $K^{th}$ index token from the codebook.

While these discrete tokens excel at capturing fine-grained low-level details~\citep{li2021token}, they lack spatial information about various features or objects within the image~\citep{darcet2023vision}. Transformers, known for their permutational-equivariant nature, efficiently model global representation~\citep{xu20232, yun2019transformers}. To incorporate local spatial information, we add learnable positional encoding $\mathbf{E}_{\text{pos}}$ to the original input (see Equation~\ref{eq:lab9}). During training, these embeddings converge into vector spaces that represent the spatial location of different tokens.

Following this spatial encoding step, the output is first processed with layer-normalization (LN) within the residual block. This helps in enhancing gradient flow and eliminates the need for an additional warm-up strategy as recommended in~\citet{xiong2020layer}.  Subsequently, the output undergoes processing via multi-head self-attention (MSA) and a multi-layer perceptron (MLP) (see Equation~\ref{eq:3}). This series of operations is repeated for a total of $L$ blocks,

\begin{equation}
\begin{aligned}
&\mathbf{out} = \mathbf{out} + \text{MSA}({\text{LN}(\mathbf{out})}), \\
& \mathbf{out} = \mathbf{out} + \text{MLP}(\text{LN}(\mathbf{out})),
\quad \Bigg\} \times L \\ 
& h_{t} = \mathbf{out}[0], \quad \texttt{MEM}_t = \mathbf{out}[0] \,.
\end{aligned}
\label{eq:3}
\end{equation}

Following $L$ blocks of operations, the feature vector associated with the \texttt{CLS} token serves as the representation, modeling both the current state and memory. This representation $h_t$ is used by the policy to sample action $\hat{a}_t \sim p_\theta\left(\hat{a}_t \mid \hat{h}_{t}\right)$ and by the critic to estimate the expected return, $v_{\xi}\left(\hat{h}_t\right) \approx \mathbb{E}_{p_\theta}\left[\sum\nolimits_{\tau \geq t} \hat{\gamma}^{\tau-t} \hat{r}_\tau\right]$. This is followed by the reward prediction, episode end prediction, and the token predictions of the next observation by the world model.

The feature vector $h_t$ now becomes the memory unit. This is possible because the self-attention mechanism acts like a gate, passing on information to the next time step as required by the task. This simple approach enables effective memory modeling without relying on recurrent networks, which can be challenging to train and struggle with long context~\citep{pascanu2013difficulty}. 

The imagination process unfolds for a duration of $H$ steps, stopping on episode-end prediction. To optimize the policy we follow a similar objective function as \gls{iris} and DreamerV2 approaches.

\section{Experiments}
We evaluated our model alongside existing baselines using the Atari 100k benchmark~\citep{kaiser2019model}, a commonly used testbed for assessing the sample-efficiency of \gls{rl} algorithms. It consists of 26 games from the Arcade Learning Environment~\citep{bellemare2013arcade}, each with distinct settings requiring perception, planning, and control skills.  

\begin{table*}[!ht]
    \centering
    \small
    \caption{DART achieves a new state-of-art median score among no-look-ahead search methods. It attains the highest median score, interquartile mean (IQM), and optimality gap score. Moreover, DART outperforms humans in 9 out of 26 games and achieves a higher score than IRIS in 18 out of 26 games (underlined).}
    \begin{tabular}{|l|ll|ll|lll|}
        \hline
        & & & \multicolumn{5}{c|}{No look-ahead search} \\
        \cline{2-8} 
         & & & & & \multicolumn{3}{c|}{Transformer based} \\
        \cline{6-8}
        Game & Random & Human & SPR & DreamerV3 & TWM & IRIS & DART \\
        \hline
        Alien & 227.8 & 7127.7 & 841.9 & 959 & 674.6 & 420.0 & \underline{\textbf{962.0}} \\
        Amidar & 5.8 & 1719.5  & \textbf{179.7} & 139 & 121.8 & 143.0 & 125.7 \\
        Assault & 222.4 & 742.0 & 565.6 & 706 & 682.6 & \textbf{1524.4} & 1316.0 \\
        Asterix & 210.0 & 8503.3 & 962.5 & 932 & \textbf{1116.6} & 853.6 & \underline{956.2} \\
        BankHeist & 14.2 & 753.1 & 345.4 & \textbf{649} & 466.7 & 53.1 & \underline{629.7} \\
        BattleZone & 2360.0 & 37187.5 & 14834.1 & 12250 & 5068.0 & 13074.0 & \underline{\textbf{15325.0}} \\
        Boxing & 0.1 & 12.1 & 35.7 & 78 & 77.5 & 70.1 & \underline{\textbf{83.0}} \\
        Breakout & 1.7 & 30.5 & 19.6 & 31 & 20.0 & \textbf{83.7} & 41.9 \\
        ChopperCommand & 811.0 & 7387.8 & 946.3 & 420 & \textbf{1697.4} & 1565.0 & 1263.8 \\
        CrazyClimber & 10780.5 & 35829.4 & 36700.5 & \textbf{97190} & 71820.4 & 59324.2 & 34070.6 \\
        DemonAttack & 152.1 & 1971.0 & 517.6 & 303 & 350.2 & 2034.4 & \underline{\textbf{2452.3}} \\
        Freeway & 0.0 & 29.6 & 19.3 & 0 & 24.3 & 31.1 & \underline{\textbf{32.2}} \\
        Frostbite & 65.2 & 4334.7 & 1170.7 & 909 & \textbf{1475.6} & 259.1 & \underline{346.8} \\
        Gopher & 257.6 & 2412.5 & 660.6 & \textbf{3730} & 1674.8 & 2236.1 & 1980.5 \\
        Hero & 1027.0 & 30826.4 & 5858.6 & \textbf{11161} & 7254.0 & 7037.4 & 4927.0 \\
        Jamesbond & 29.0 & 302.8 & 366.5 & 445 & 362.4 & \textbf{462.7} & 353.1 \\
        Kangaroo & 52.0 & 3035.0 & 3617.4 & \textbf{4098} & 1240.0 & 838.2 & \underline{2380.0} \\
        Krull & 1598.0 & 2665.5 & 3681.6 & \textbf{7782} & 6349.2 & 6616.4 & \underline{7658.3} \\
        KungFuMaster & 258.5 & 22736.3 & 14783.2 & 21420 & \textbf{24554.6} & 21759.8 & \underline{23744.3}\\
        MsPacman & 307.3 & 6951.6 & 1318.4 & 1327 & \textbf{1588.4} & 999.1 &  \underline{1132.7}\\
        Pong & -20.7 & 14.6 & -5.4 & 18 & \textbf{18.8} & 14.6 & \underline{17.2}\\
        PrivateEye & 24.9 & 69571.3 & 86.0 & \textbf{882} & 86.6 & 100.0 & \underline{765.7}\\
        Qbert & 163.9 & 13455.0 & 866.3 & \textbf{3405} & 3330.8 & 745.7 & \underline{750.9}\\
        RoadRunner & 11.5 & 7845.0 & 12213.1 & \textbf{15565} & 9109.0 & 4046.2  & \underline{7772.5}\\
        Seaquest & 68.4 & 42054.7 & 558.1 & 618 & 774.4 & 661.3 & \underline{\textbf{895.8}}\\
        UpNDown & 533.4 & 11693.2 & 10859.2 & 7667 & \textbf{15981.7} & 3546.2  & \underline{3954.5}\\       
        \hline
        \#Superhuman($\uparrow$) & 0 & N/A & 6 & 9 & 7 & 9 & 9\\
        Mean($\uparrow$) & 0.000 & 1.000 & 0.616 & 1.120 & 0.956 & 1.046  & 1.022\\
        Median($\uparrow$) & 0.000 & 1.000 & 0.396 & 0.466 & 0.505 & 0.289 & \underline{\textbf{0.790}}\\
        IQM($\uparrow$) & 0.000 & 1.000 & 0.337 & 0.490 & - & 0.501 & \underline{\textbf{0.575}}\\
        Optimality Gap($\downarrow$) & 1.000 & 0.000 & 0.577 & 0.508 & - & 0.512 &  \underline{\textbf{0.458}}\\
        \hline
    \end{tabular}
    \label{tab:1}
\end{table*}

We evaluated our model's performance based on several metrics, including the mean and median of the human-normalized score, which measures how well the agent performs compared to human and random players given as $\frac{{\text{score}_{\text{agent}} - \text{score}_{\text{random}}}}{{\text{score}_{\text{human}} - \text{score}_{\text{random}}}}$. We also used the super-human score to quantify the number of games in which our model outperformed human players. We further evaluated our model's performance using the Interquartile Mean (IQM) score and the Optimality Gap, following the evaluation guidelines outlined in~\citet{agarwal2021deep}.

We rely on the median score to evaluate overall model performance, as it is less affected by outliers. The mean score can be strongly influenced by a few games with exceptional or poor performance. Additionally, the IQM score helps in assessing both consistency and average performance across all games.

Atari environments offer the model an RGB observation of $64 \times 64$ dimensions, featuring a discrete action space, and the model is allowed to be trained using only 100k environment steps (equivalent to 400k frames due to a frameskip of 4), which translates to approximately 2 hours of real-time gameplay.

The world model is trained with a GPT-style causal (decoder) transformer, while the policy is trained using a ViT-style (encoder) transformer. This allows for parallel computation of multiple steps during world model training, making it computationally much faster than previous methods like the recurrent network-based DreamerV3. During policy training, actions for each time step are computed using the modeled CLS token, which then serves as a memory token for the next step. Although this process is computed step by step, it remains efficient compared to methods like IRIS, as it doesn't require additional networks like LSTM to retain memory.

\subsection{Results} 
\begin{figure*}[htbp]
    \centering
    \includegraphics[width=\linewidth]{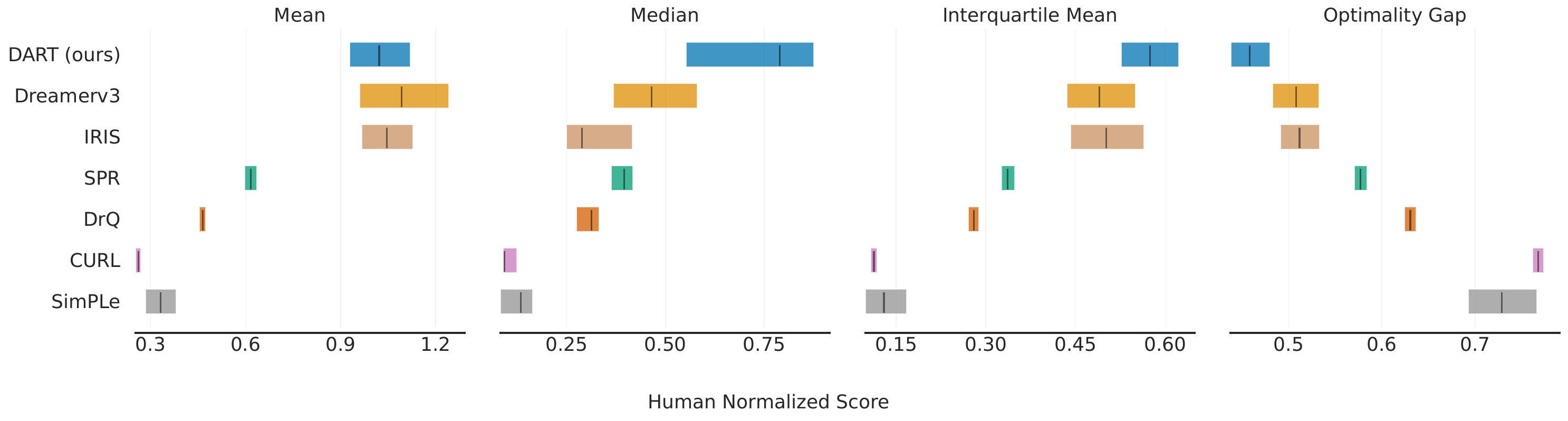}
    \caption{Comparison of Mean, Median, and Interquartile Mean Human-Normalized Scores}
    \label{fig:comparison_scores}
\end{figure*}

\begin{figure*}[!ht]
    \centering
    \begin{subfigure}{0.48\linewidth}
        \centering
        \includegraphics[width=\linewidth]{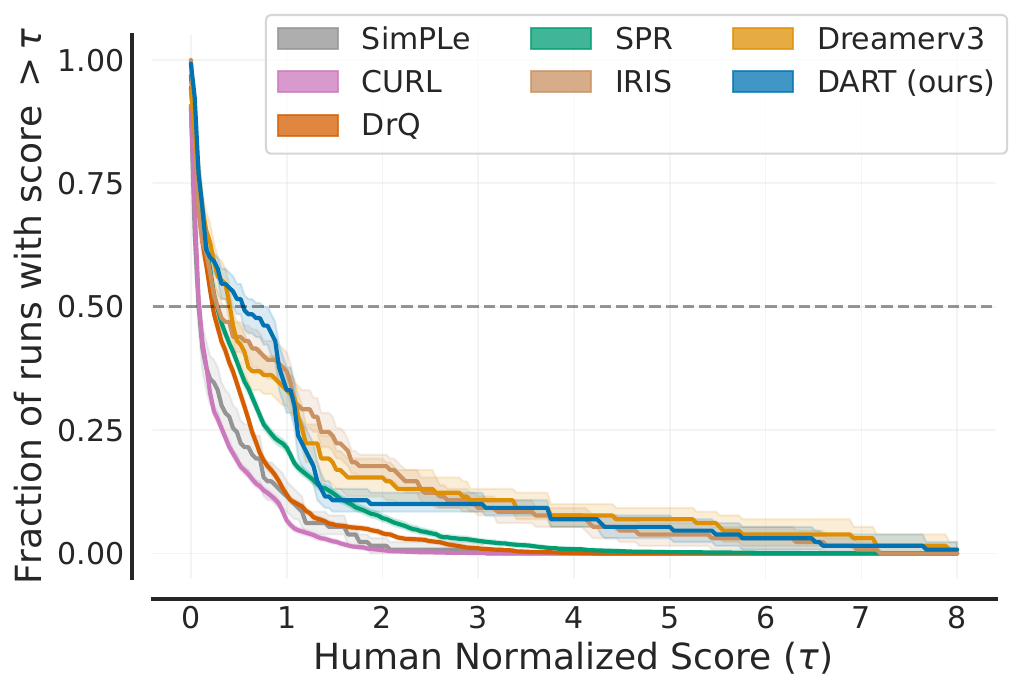}
        \caption{The performance profiles on the Atari 100k benchmark illustrate the proportion of runs across all games (y-axis) that achieve a score normalized against human performance (x-axis).}
        \label{fig:subfig1}
    \end{subfigure}
    \hfill
    \begin{subfigure}{0.48\linewidth}
        \centering
        \includegraphics[width=\linewidth]{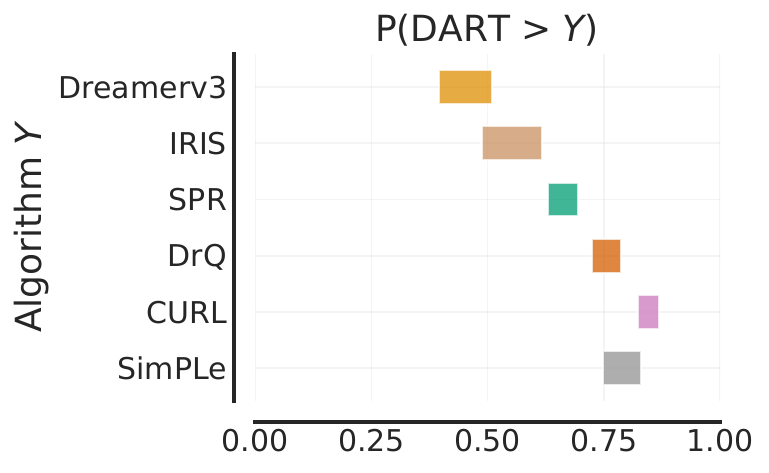}
        \caption{The probabilities of improvement visualized here refer to the likelihood of \gls{dart} surpassing the performance of baseline models in any game.}
        \label{fig:subfig2}
    \end{subfigure}
    \caption{Comparison of different models using performance profiles and probabilities of improvement.}
    \label{fig:main}
\end{figure*}

In Figure~\ref{fig:comparison_scores}, we present the IQM and optimality gap scores, as well as the mean and median scores. These scores pertain to various models assessed on Atari 100k. Figure~\ref{fig:subfig1} visualizes the performance profile, while Figure~\ref{fig:subfig2} illustrates the probability of improvement, which quantifies the likelihood of \gls{dart} surpassing baseline models in any Atari game. To perform these comparisons, we use results from~\citet{micheli2022transformers}, which include scores of 100 runs of CURL~\citep{laskin2020curl}, DrQ~\citep{kostrikov2020image}, SPR~\citep{schwarzer2020data}, as well as data from 5 runs of SimPLe~\citep{kaiser2019model} and \gls{iris}.  

\gls{dart} exhibits a similar mean performance as \gls{iris}. However, the median and IQM scores show that \gls{dart} outperforms other models consistently. 

Table~\ref{tab:1} presents \gls{dart}'s score across all 26 games featured in the Atari 100k benchmark. We compare its performance against other strong world models including DreamerV3~\citep{hafner2023mastering}, as well as other transformer-based world models, such as \gls{twm}~\citep{robine2023transformer} and \gls{iris}~\citep{micheli2022transformers}.

To assess \gls{dart}'s overall performance, we calculate the average score over 100 episodes post-training, utilizing five different seeds to ensure robustness. \gls{dart} outperforms the previous best model, \gls{iris}, in 18 out of 26 games. It achieves a median score of 0.790 (an improvement of 61\% when compared to DreamerV3). Additionally, it reaches an IQM of  0.575 reflecting a 15\% advancement, and significantly improves the OG score to 0.458, indicating a 10\% improvement when compared to IRIS. \gls{dart} also achieves a superhuman score of 9, outperforming humans in 9 out of 26 games.

\subsection{Policy Analysis} In Figure~\ref{fig:four}, we present the attention maps for the 6 layers of our transformer policy using a heat-map visualization. These maps are generated by averaging the attention scores from each multi-head attention mechanism across all layers. The final visualization is obtained by further averaging these attention maps over 20 randomly selected observation states during an episode. This analysis provides insights into our approach to information aggregation through self-attention.

The visualization in Figure~\ref{fig:four} shows that the extent to which information is aggregated from the past and the current state to the next state depends on the specific task at hand. In games featuring slowly moving objects where the current observation provides complete information to the agent, the memory token receives less attention (see Figure~\ref{fig:subfig4}). Conversely, in environments with fast-moving objects like balls and paddles, where the agent needs to model the past trajectory of objects (e.g., Breakout and Private Eye), the memory token is given more attention (see Figure~\ref{fig:subfig3}-~\ref{fig:subfig6}). This observation highlights the adaptability of our approach to varying task requirements.

On further analysis, we observe that DART performs better in environments with many approaching enemies and infraction, such as Alien (three enemies need to be tracked) and Seaquest (keep track of divers and dodge enemy subs and killer sharks). However, DreamerV3 does better in games where global information is enough for planning actions. This is because DART uses discrete tokens to focus on important task-related details with its attention mechanism, while DreamerV3's approach suits games with fewer components. We saw a similar pattern in long, complex tasks with multiple infractions in the game of Crafter (Section \ref{Sec:A}), where DART showed improved performance over DreamerV3 in modeling long horizon tasks with multiple components.

\subsection{Ablation Studies} We further analyzed \glspl{dart} performance across various experimental settings, as detailed in Table~\ref{tab:5} for five distinct games. The original score of \gls{dart} is presented in the second column. The different scenarios include:

\textbf{Without Positional Encoding (PE):} The third column demonstrates the performance of \gls{dart} when learned positional encoding is excluded. We can observe that in environments where agents need to closely interact with their surroundings, such as in Boxing and KungFuMaster, the omission of positional encoding significantly impacts performance. However, in games where the enemy may not be in close proximity to the agent, such as Amidar, there is a slight drop in performance without positional encoding. This is because transformers inherently model global context, allowing the agent to plan its actions based on knowledge of the overall environment state. However, precise decision-making requires positional information about the local context. In our case, adding learnable positional encoding provides this, resulting in a significant performance boost.

\textbf{No Exploration ($\epsilon$):} The fourth column illustrates \glspl{dart} performance when trained without random exploration, relying solely on agent-predicted actions for collecting trajectories for world modeling. However, like \gls{iris}, our model also faces the double-exploration challenge. This means that the agent's performance declines when new environment states aren't introduced through random exploration, which is crucial for effectively modeling the dynamics of the world. It's worth noting that for environments with simpler dynamics (e.g., Seaquest), the performance impact isn't as substantial.

\textbf{Masking Memory Tokens:} In the fifth column, we explore the impact of masking the memory token, thereby removing past information. Proper modeling of memory is crucial in \gls{rl} to address the challenge of partial observability and provide information about various states (e.g., the approaching trajectory of a ball, and the velocity of the surrounding objects) that are important for decision-making. Our method of aggregating memory over time enhances \glspl{dart} overall performance. However, since Atari games exhibit diverse dynamics, the effect of masking the memory tokens varies accordingly.

\begin{figure*}[!ht]
    \centering
       \begin{subfigure}{0.45\linewidth}
        \centering
        \includegraphics[width=\linewidth]{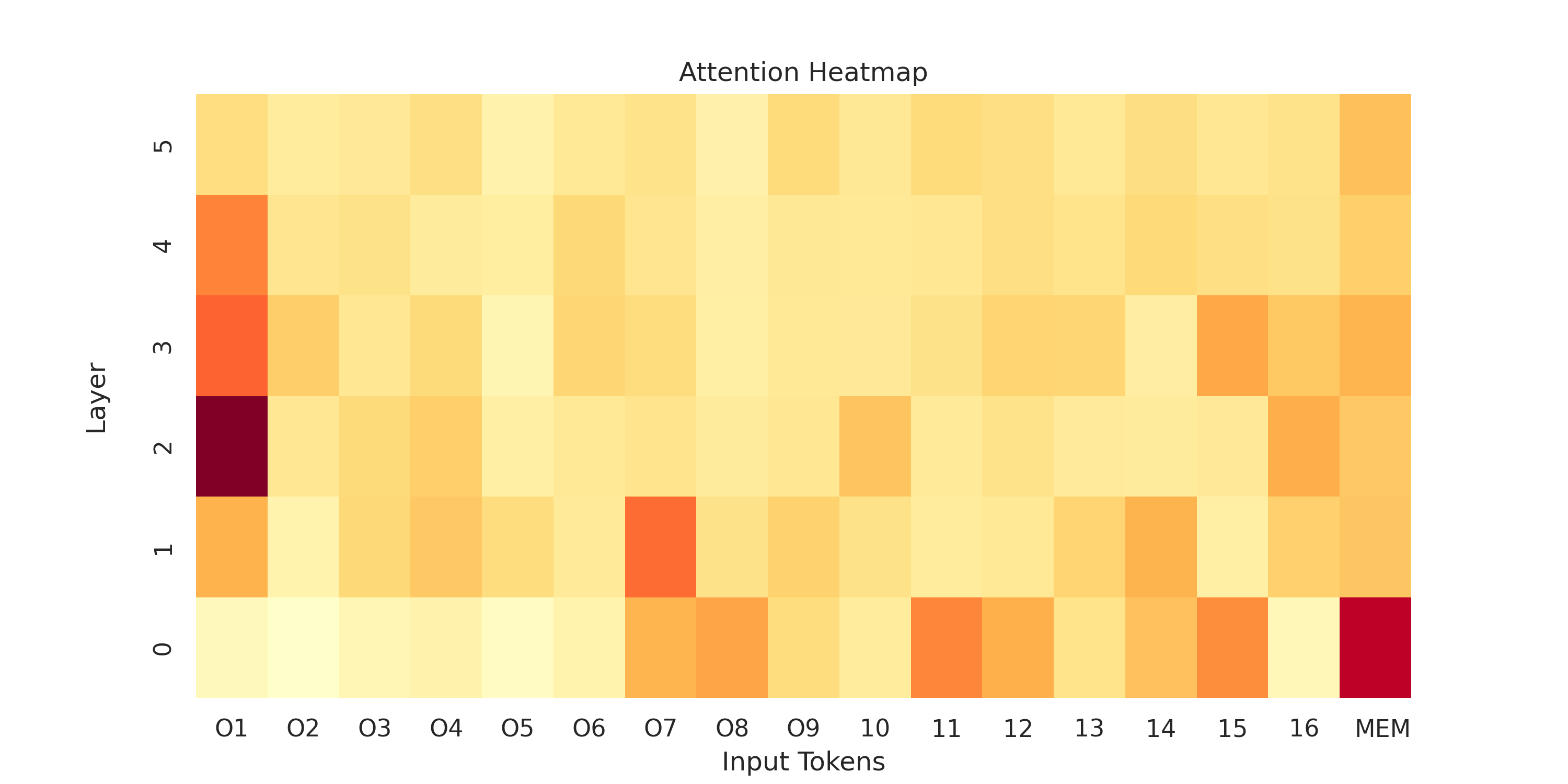}
        \caption{Amidar}
        \label{fig:subfig4}
    \end{subfigure}
    \begin{subfigure}{0.45\linewidth}
        \centering
        \includegraphics[width=\linewidth]{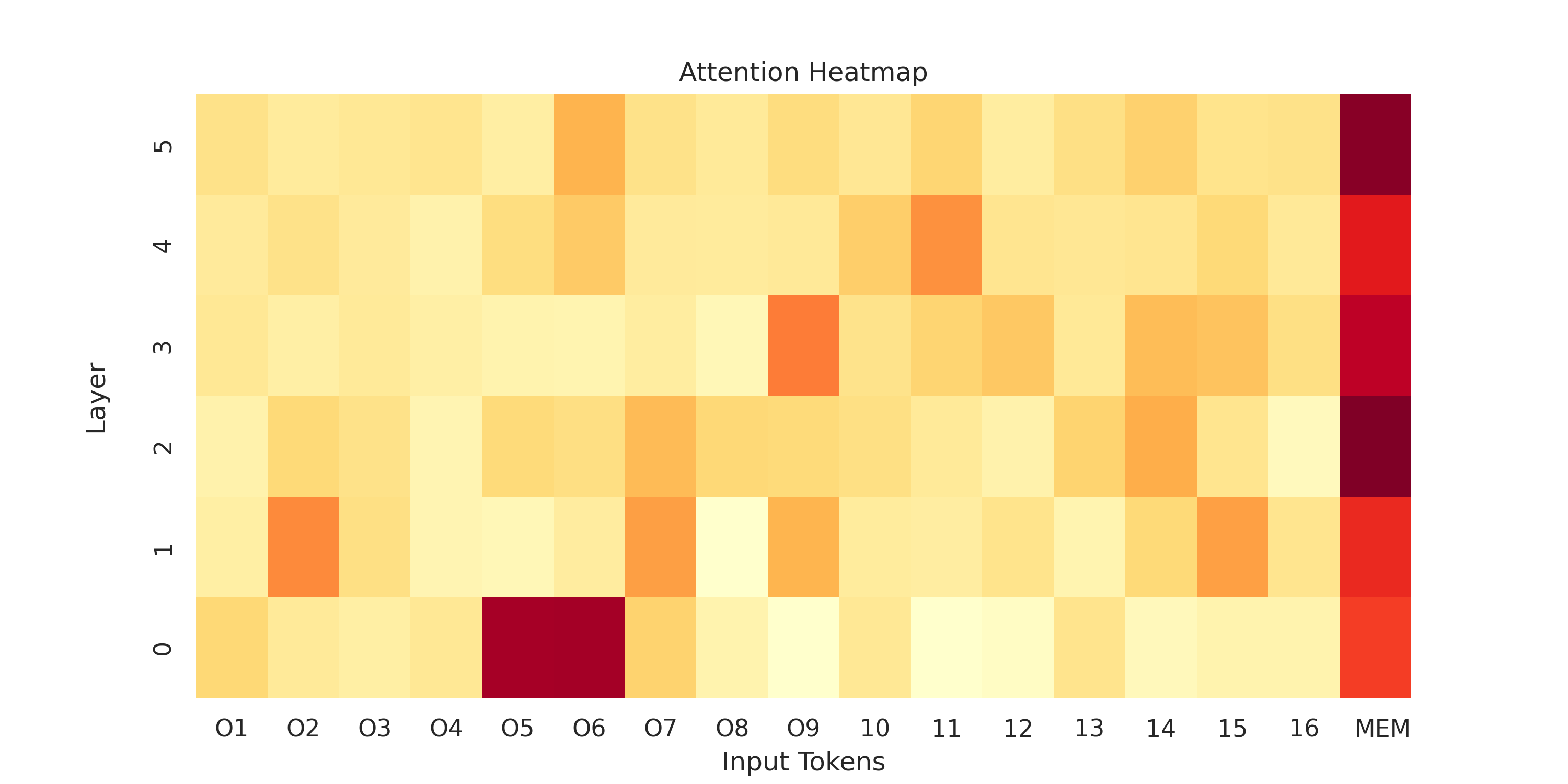}
        \caption{Breakout}
        \label{fig:subfig3}
    \end{subfigure}
    \vspace{0.5cm} % Adjust the vertical space between rows
    
    \begin{subfigure}{0.45\linewidth}
        \centering
        \includegraphics[width=\linewidth]{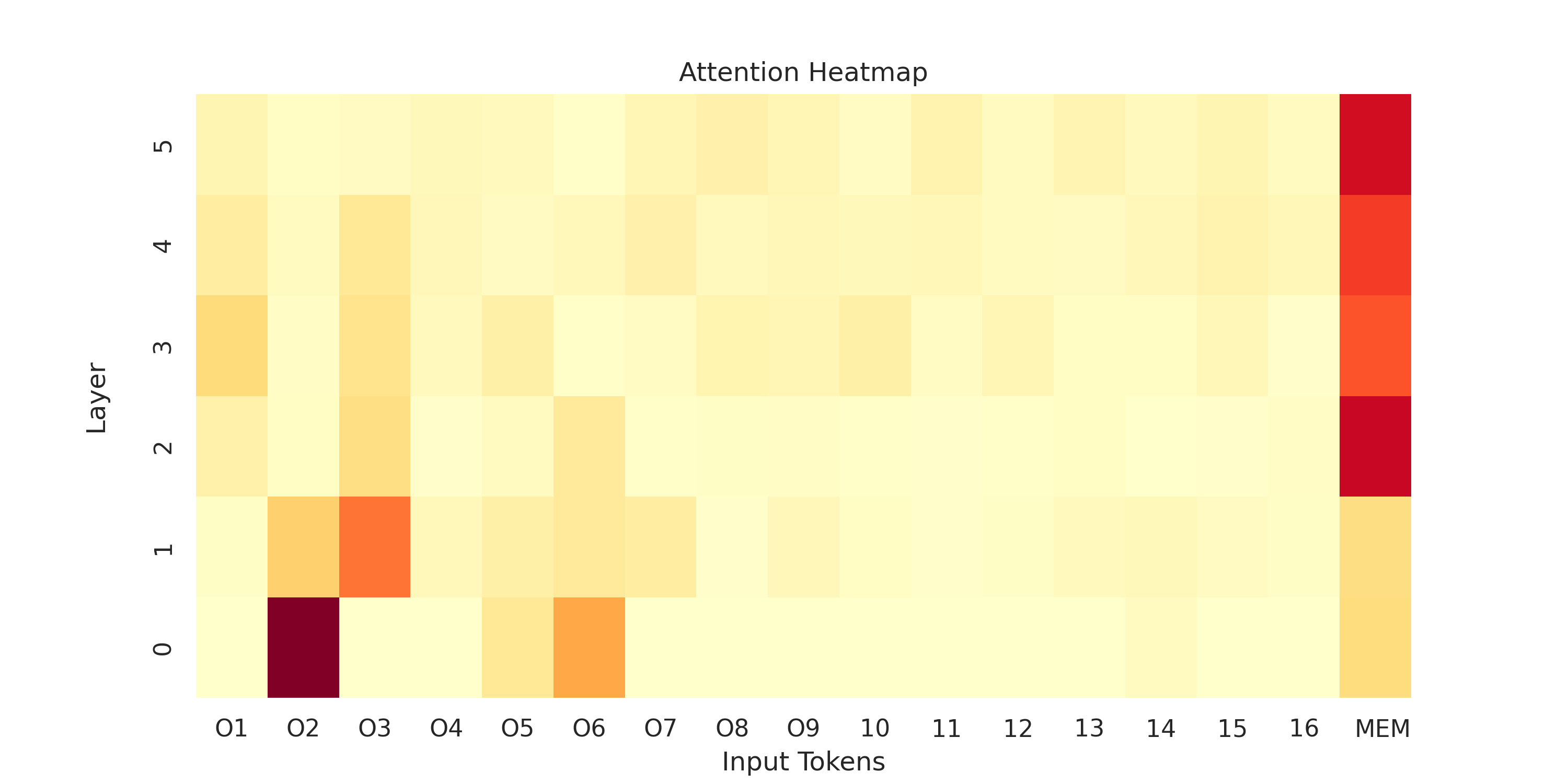}
        \caption{Private Eye}
        \label{fig:subfig5}
    \end{subfigure}
    \begin{subfigure}{0.45\linewidth}
        \centering
        \includegraphics[width=\linewidth]{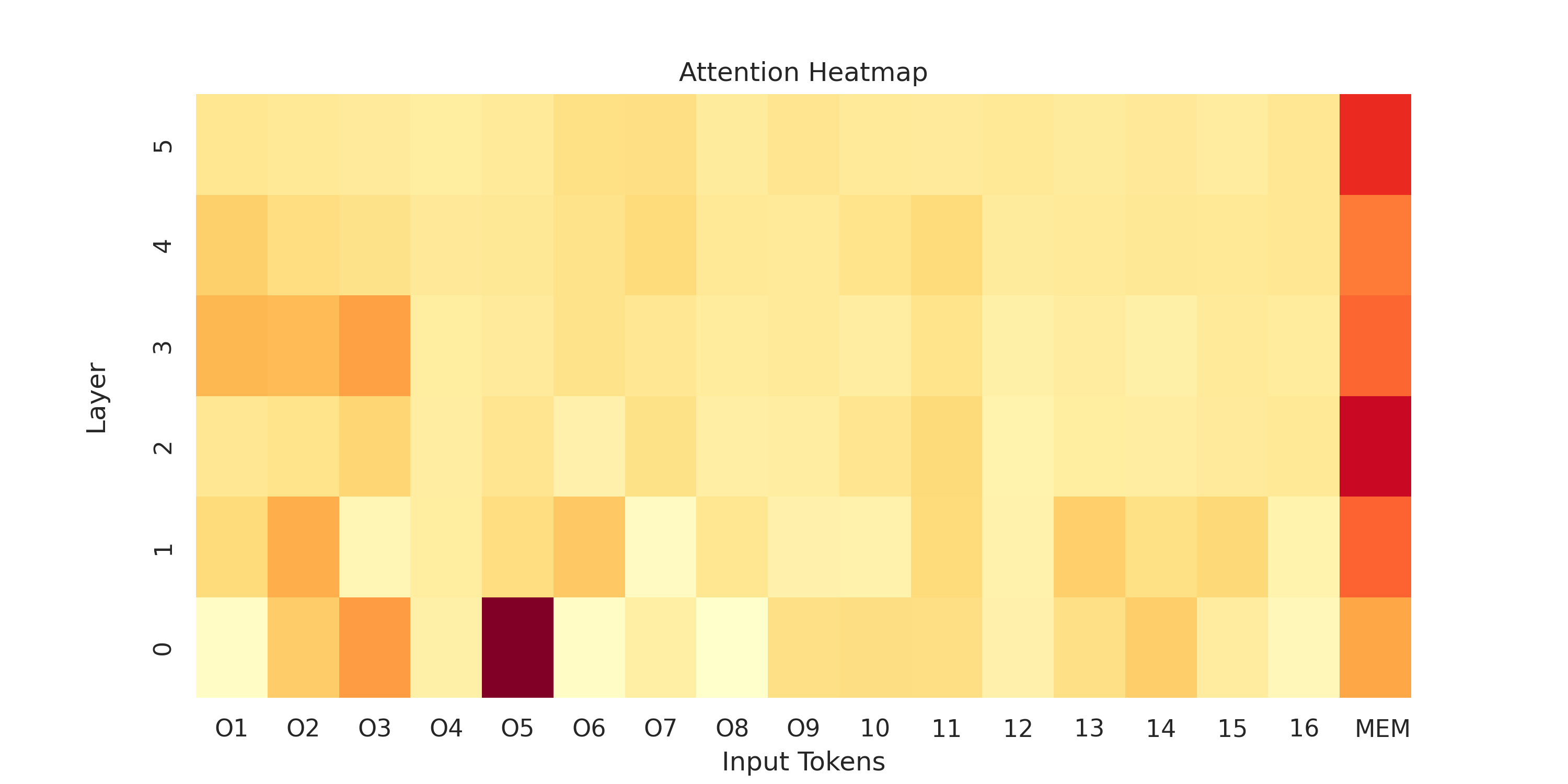}
        \caption{Krull}
        \label{fig:subfig6}
    \end{subfigure}
    
    \caption{ \textbf{Comparison of Memory Requirements Across Atari Games:} Atari games exhibit varying memory requirements, depending on their specific dynamics. Games with relatively static or slow-moving objects, like \textit{Amidar}, maintain complete information at each time step and thus aggregate less information from the memory token. Conversely, games characterized by rapidly changing environments, such as \textit{Breakout}, \textit{Krull}, and \textit{PrivateEye}, require modeling the past trajectories of objects. As a result, the policy for these games heavily relies on the memory token to aggregate information from past states into future states.}
    \label{fig:four}
\end{figure*}

In some games, decisions are solely based on information from the current time step, making memory tokens unnecessary. However, in other games, such as Breakout, Private eye, and Krull, tracking memory is crucial for optimal planning, like predicting the ball's trajectory or following clues. This is evident in the heatmap visualization in Figure~\ref{fig:four}, where significant attention is given to memory tokens for the aforementioned games. On the contrary, in games like Boxing and Amidar, where long-term trajectory information or extensive planning isn't needed, relying solely on recent state information is often sufficient for optimal decision-making, and thus there is only a small impact on the final performance with the masking of memory tokens.

It is interesting to observe improvement in the agent's performance with masked memory tokens in the case of RoadRunner. This could be because the original state already contains complete information, rendering the memory token redundant, thereby impacting the final performance.

\textbf{Random Observation Token Masking:} The last set of columns explores the consequences of randomly masking observation tokens, which selectively removes low-level information. Given that each token among the $K$ tokens model distinct low-level features of the observation, random masking has a noticeable impact on the agent's final performance. When observation tokens are masked 100\%, the agent attends solely to the memory token, resulting in a significant drop in overall performance.
\begin{table*}[!ht]
    \centering
    \small
    \caption{Evaluating DART's performance through various techniques such as memory token masking, random observation masking, and the removal of positional encoding and random exploration.}
    
    \begin{tabular}{lllllllll}
        \hline
        & & & & & \multicolumn{4}{c}{Masked} \\
        & & \multicolumn{2}{c}{w/o} & Masked &  \multicolumn{4}{c}{Observation Token} \\
        \cline{3-4} \cline{6-9}
        Game & Original & PE & $\epsilon$ & Memory & 25\%  & 50\%  & 75\%  & 100\% \\
        \hline
        Boxing  &  83.0 & 3.86 & 58.67 & 81.45 & 77.79 &  51.14 & 15.64 & -11.91 \\
        Amidar  & 125.7  & 77.1 & 92.75 & 113.69 &  
 102.47 & 56.37 & 52.22 & 30.43 \\
        Road Runner  & 7772.5 & 1030.0  & 3597.1 & 8021.0 & 7354.0  & 2730.0 & 988.0 & 961.0 \\
        Seaquest  & 895.8 & 64.2 & 753.93 & 704.8 & 491.4 & 207.8 &  104.0 & 142.0 \\
        KungFuMaster  & 23744.3 & 1028.0 & 15464.7 & 20378.0& 16436.0 &  9760.0 & 4676.2 & 1571.8  \\       
        \hline
    \end{tabular}
    \label{tab:5}
\end{table*}

\section{Related Work}

\noindent \textbf{Sample Efficiency in RL.} Enhancing sample efficiency (i.e., the amount of data required to reach a specific performance level) constitutes a fundamental challenge in the field of \gls{rl}. This efficiency directly impacts the time and resources needed for training an \gls{rl} agent. Numerous approaches aimed at accelerating the learning process of \gls{rl} agents have been proposed~\citep{buckman2018sample, mai2022sample, yu2018towards}. Model-based \gls{rl} is one such approach that helps improve the sample efficiency. It reduces the number of interactions an agent needs to have with the environment to learn the policy~\citep{moerland2023model, polydoros2017survey, atkeson1997comparison}. This is done by allowing the policy to learn the task in the imagined world~\citep{wang2021offline, mu2021model, okada2021dreaming, zhu2020bridging}. This motivates the need to have an accurate world model while providing the agent with concise and meaningful task-relevant information for faster learning. Considering this challenge~\citet{kurutach2018model} learns an ensemble of models to reduce the impact of model bias and variance. Uncertainty estimation is another approach as shown in~\citet{plaat2023high} to improve model accuracy. It involves estimating the uncertainty in the model's prediction so that the agent focuses its exploration in those areas. The other most common approach for an accurate world model is using a complex or higher-capacity model architecture that is better suited to the task at hand~\citep{wang2021ed2, ji2022update}. For example, using a transformer-based world model, as in TransDreamer~\citep{chen2022transdreamer}, \gls{twm}~\citep{robine2023transformer}, and \gls{iris}~\citep{micheli2022transformers}.

Learning a low-dimensional representation of the environment can also help improve the sample efficiency of \gls{rl} agents. By reducing the dimensionality of the state, the agent can learn an accurate policy with fewer interactions with the environment~\citep{mcinroe2021learning, du2019good}. Variational Autoencoders (VAEs)~\citep{kingma2019introduction} are commonly used for learning low-dimensional representations in \gls{mbrl}~\citep{andersen2018dreaming}. The VAEs capture a compact and informative representation of the input data. This allows the agent to learn the policy faster~\citep{ke2018modeling, corneil2018efficient}. However, VAEs learn a continuous representation of the input data by forcing the latent variable to be normally distributed. This poses a challenge for \gls{rl} agents, where agents need to focus on precise details for decision-making~\citep{dunion2022temporal}. \citet{lee2020weakly} show disentangling representations helps in modeling interpretable policy and improves the learning speed of \gls{rl} agents on various manipulation tasks. Recent works~\citep{robine2023smaller, zhang2022deep} have used \gls{vq-vae} for learning independent latent representations of different low-level features present in the original observation. Their clustering properties have enabled robust, interpretable, and generalizable policy across a wide range of tasks.

\section{Conclusion}
In this work, we introduced \gls{dart}, a model-based reinforcement learning agent that learns both the model and the policy using discrete tokens. Through our experiments, we demonstrated our approach helps in improving performance and achieves a new state-of-the-art score on the Atari 100k benchmarks for methods with no look-ahead search during inference. Moreover, our approach for memory modeling and the use of a transformer for policy modeling provide additional benefits in terms of interpretability. 

\noindent \textbf{Limitations:} As of now, our method is primarily designed for environments with discrete action spaces. This limitation poses a significant challenge, considering that many real-world robotic control tasks necessitate continuous action spaces. For future work, it would be interesting to adapt our approach to continuous action spaces and modeling better-disentangled tokens for faster learning. 

\section*{Acknowledgement}
The authors would like to thank the reviewers for their valuable feedback. We are also thankful to the Digital Research Alliance of Canada for the computing resources and CIFAR for research funding.

\section*{Impact Statement}
This paper presents works where the goal is to advance the field of Machine Learning. There are many potential societal consequences of our work, none of which we feel must be specifically highlighted here.

\bibliography{main}
\bibliographystyle{icml2024}

\newpage
\appendix
\onecolumn
\section{Appendix}

\subsection{Experiment on Crafter}
\label{Sec:A}

Crafter~\citep{hafner2021benchmarking}, inspired by Minecraft~\citep{guss2019minerl}, allows assessing an agent's general abilities within a single environment. This distinguishes it from Atari 100k, where the agent must be evaluated across 26 different games that test for different skills. In Crafter, 2D worlds are randomly generated, featuring diverse landscapes like forests, lakes, mountains, and caves on a 64$\times$64 grid. Players aim to survive by searching for essentials like food, water, and shelter while defending against monsters, collecting materials, and crafting tools. This setup allows for evaluating a wide range of skills within a single environment, spanning multiple domains, and increasing assessment comprehensiveness.
The environment is partially observable with observations covering a small 9$\times$9 region centered around the agent.

\begin{table}[h]
    \centering
    \caption{Comparing the sample efficiency of DreameV3, IRIS, and DART on challenging Crafter environment which involves long-horizon tasks. Reported returns are specified as average and standard deviation over 5 seeds.}
    \begin{tabular}{|c|ccc|}
        \hline
        Model & DreamerV3 & IRIS & DART \\
        \hline
        Steps & 1M & 1M & 1M\\
        \hline
        Return & $11.7 \pm 1.9$ & $9.23\pm0.56$ & {$\mathbf{12.2\pm1.67}$} \\
        \hline
    \end{tabular}
    \label{tab:2}
\end{table}

In the results shown in Table~\ref{tab:2}, we compare DART with IRIS and Dreamer V3 in a low data regime and observed that DART achieves a higher average return, further showcasing the efficiency of DART over previous models.

\subsection{Experiment on Atari with More Environment Steps}

\begin{table}[h]
    \centering
    \caption{Performance of DART with 100k and 150k environment steps (\textit{k}). All results are shown as average and standard deviation over 5 seeds.}
    \begin{tabular}{|l|c|c|}
    \hline
    \multirow{2}{*}{\textbf{Environment}} & \textbf{Steps} & \textbf{Score} \\
    & (\textit{k}) & \\
    \hline
    \multirow{2}{*}{Freeway}  & 100k & 32.2 $\pm$ 0.57 \\
    & 150k & 33.1 $\pm$ 0.37 \\
    \hline
    \multirow{2}{*}{KungFuMaster} & 100k & 23744.3 $\pm$ 3271.53\\
    & 150k & 24756.5 $\pm$ 2635.21\\
    \hline
    \multirow{2}{*}{Pong} & 100k & 17.2 $\pm$ 1.74 \\
    & 150k & 17.6 $\pm$ 2.79  \\
    \hline
    \end{tabular}
    \label{tab:10}
\end{table}
By training it beyond 100k training steps, we see improved performance of \gls{dart} as shown in Table~\ref{tab:10}, showcasing the scalability of \gls{dart} with more data.

\subsection{Model Configuration}
Recent works have used transformer-based architectures for \gls{mbrl}. In Table~\ref{tab:9} we compare the configurations used by different approaches for representation learning, world modeling, and behavior learning.   

\begin{table}[!ht]
    \centering
    \caption{Comparing the model configuration of recent \gls{mbrl} approaches. n/a- Not Available; Cat.-VAE - Categorical VAE.; MAE - Masked Auto Encoder }
    \small
    \begin{tabular}{lllllll}
    \hline
    & MWM & TWM & IRIS & DreamerV3 & STORM & DART \\
    \hline
    Parameters &n/a&n/a& 3.04M & 18M & n/a & 3.07M \\
    State model & MLP & MLP & CNN & MLP & MLP & ViT \\
    Agent memory & ViT & Tr-XL & LSTM & GRU & GPT & ViT (Self-attention) \\
    Representation & MAE & Cat.-VAE & VQ-VAE & Cat.-VAE & Cat.-VAE & VQ-VAE \\

    \hline
    \end{tabular}
    \label{tab:9}
\end{table}

\subsection{Hyperparameters}
A detailed list of hyperparameters is provided for each module: Table~\ref{tab:6} for Image Tokenizer, Table~\ref{tab:7} for World Modeling, and Table~\ref{tab:8} for behaviour learning.

\begin{table}[!ht]
    \centering
    \small
    \caption{Hyperparameters for image tokenization using VQ-VAE.}
    \begin{tabular}{|l|l|l|}
    \hline
    \textbf{Hyperparameter} & \textbf{Symbol} & \textbf{Value} \\
    \hline
    Encoder convolutional layers & -- & 4 \\
    Decoder convolutional layers & -- & 4 \\
    Per layer residual blocks & -- & 2 \\
    Self-attention layers & -- & 8 / 16 \\
    Codebook size & $N$ & 512 \\
    Embedding dimension & $d$ & 512 \\
    Input image resolution & -- & 64$\times$64 \\
    Image channels & -- & 3 \\
    Activation & -- & Swish \\
    Tokens per image & $K$ & 16 \\
    Batch size & -- & 64 \\
    Learning rate & -- & 0.0001\\
    \hline
    \end{tabular}
    \label{tab:6}
\end{table}

\begin{table}[!ht]
    \centering
    \small
    \caption{Hyperparameters used for modeling the dynamics using transformer decoder.}
    
    \begin{tabular}{|l|l|l|}
    \hline
    \textbf{Hyperparameter} & \textbf{Symbol} & \textbf{Value} \\
    \hline
    Embedding dimension & -- & 256 \\
    Transformer layers & -- &  10 \\
    Attention heads & -- & 4 
\\
    Imagination steps & $H$ & 20 \\
    Embedding dropout & -- & 0.1 \\
    Weight decay & -- & 0.01 \\
    Attention dropout & -- & 0.1 \\
    Residual dropout & -- & 0.1 \\   
    Attention type & -- & Causal \\
    Activation & -- & GeLU \\
    Batch size & -- & 64 \\
    Learning rate & -- & 0.0001\\
    \hline
    \end{tabular}
    \label{tab:7}
\end{table}

\begin{table}[!ht]
    \centering
    \small
    \caption{Hyperparameters used for modeling behavior using transformer encoder.}
    
    \begin{tabular}{|l|l|l|}
    \hline
    \textbf{Hyperparameter} & \textbf{Symbol} & \textbf{Value} \\
    \hline
    Input tokens & -- & 18 \\
    Embedding dimension & -- & 512 \\
    Attention heads & -- & 8 \\
    Transformer layers & $L$ & 6 \\
    Dropout & -- & 0.2 \\
    Activation & -- & GeLU \\
    Transformer layers &  -- & 6 \\
    Attention type & -- & Self-attention \\
    Positional embedding & -- & Learnable \\
    Gamma & $\gamma$ & 0.995 \\
    Lambda & $\lambda$ & 0.95 \\
    Batch size & -- & 64 \\
    Epsilon & $\epsilon$  & 0.01 \\
    Temperature (train) & -- & 1.0 \\
    Temperature (test) & -- & 0.5 \\
    Learning rate & -- & 0.0001\\
    \hline
    \end{tabular}
    \label{tab:8}
\end{table}

\subsection{Comparing the performance with STORM}
Recently released another transformer-based model STORM~\citep{zhang2023storm} showcases close to similar performance when compared with \gls{dart} as shown in Table~\ref{tab:15}. STORM relies on utilizing VAEs for modeling stochastic world models, while we use VQ-VAE for modeling discrete world models. While STORM performs similarly to DART overall, it struggles in games like Pong and Breakout with single small moving objects. Comparing DART's performance with STORM, there's a notable improvement in DART's performance due to its discrete representation. DART's use of discrete tokens for each entity, coupled with an attention mechanism, enables the agent to focus precisely on relevant tasks, leading to significant performance boosts, especially in such games. Additionally, this approach makes DART more interpretable compared to STORM, as illustrated in Figure~\ref{fig:four}'s heatmap visualization.  
\begin{table}[!ht]
    \centering
    \small
    \caption{Comparing the performance of DART with STORM.}
    \begin{tabular}{|l|ll|}
        \hline
        Game & STORM & DART \\
        \hline
        Alien & 984 & 962.0 \\
        Amidar & 205 & 125.7 \\
        Assault  & 801 & 1316.0 \\
        Asterix & 1028 & 956.2 \\
        BankHeist & 641 & 629.7 \\
        BattleZone & 13540 & 15325.0 \\
        Boxing & 80 & 83.0 \\
        Breakout & 16 & 41.9 \\
        ChopperCommand & 1888 & 1263.8 \\
        CrazyClimber & 66776 & 34070.6 \\
        DemonAttack & 165 & 2452.3 \\
        Freeway &0 & 32.2 \\
        Frostbite  &1316 & 346.8 \\
        Gopher & 8240 & 1980.5 \\
        Hero & 11044 & 4927.0 \\
        Jamesbond & 509 & 353.1 \\
        Kangaroo & 4208 & 2380.0 \\
        Krull &8413 & 7658.3 \\
        KungFuMaster &26182 & 23744.3\\
        MsPacman & 2673 &  1132.7\\
        Pong & 11 & 17.2\\
        PrivateEye & 7781&765.7\\
        Qbert & 4522&750.9\\
        RoadRunner & 17564& 7772.5\\
        Seaquest & 525 &895.8\\
        UpNDown & 7985 & 3954.5\\       
        \hline
        \#Superhuman($\uparrow$) & 9 & 9\\
        Mean($\uparrow$) &1.267 &  1.022\\
        Median($\uparrow$) & 0.584 & 0.790\\
        % IQM($\uparrow$) & - & 0.575\\
        % Optimality Gap($\downarrow$) & - & 0.458\\
        \hline
    \end{tabular}
    \label{tab:15}
\end{table}

\begin{algorithm}
\caption{Integrating DART with lookahead search methods}\label{alg:integrate_DART}
\begin{algorithmic}[1]
\STATE Initialize pretrained dynamics model $f$ (transformer decoder) and policy $\pi$ (transformer encoder).
\FOR{each episode}
\STATE Sample initial state $s$ from the environment
\STATE Map state $s$ into a sequence of discrete tokens $s_0,s_1,...,s_N$ using pre-trained VQ-VAE
\FOR{$t=0$ to $T-1$}
\STATE \textbf{Planning Action Sequence:}
\STATE $\quad$ Given current state tokens, use transformer decoder $f$ to simulate possible future state tokens $s'$ and rewards for each possible action
\STATE $\quad$ Apply a planning algorithm (e.g., Monte Carlo Tree Search) to search for the optimal sequence of actions that maximize cumulative reward
\STATE $\quad$ Select the first action $a_t$ from the optimal sequence
\STATE Execute $a_t$, observe next state $s_{t+1}$ and reward $r_t$
\ENDFOR
\ENDFOR
\end{algorithmic}
\end{algorithm}

\subsection{Integrating DART with lookahead search methods.}
The proposed model \gls{dart} doesn't utilize lookahead search, limiting its ability to leverage the learned dynamics for future trajectory planning. In contrast, Efficient Zero~\citep{ye2021mastering} the state-of-the-art model-based reinforcement learning uses the lookahead search method to plan future trajectories based on the learned model of the environment. Simulating future states and rewards, enables more effective decision-making. This approach predicts outcomes of different actions and selects the best course of action, utilizing \gls{mcts} for planning future trajectories.

Our method of modeling \gls{dart} with discrete representation enables it to be easily integrated with lookahead search methods. In the Algorithm~\ref{alg:integrate_DART} below, we'll briefly outline how we integrate the lookahead search method into \gls{dart} for efficient future trajectory planning, which we plan to explore further in future work. However, this approach is computationally expensive because exploring the entire state-action space is infeasible. Additionally, exploring the vast state-action space at each iteration requires significant computational resources, posing a challenge. Therefore, we currently limit DART's results without incorporating lookahead search methods.

\subsection{Modeling DART for continuous action space.}
As outlined in Algorithm~\ref{alg:DART_continuous}, adapting DART for continuous action space involves training an additional action tokenizer, alongside an image tokenizer as explained in Section~\ref{sec2.1}. The action tokenizer converts continuous action tokens into discrete codebook embeddings, while the world modeling process remains unchanged. In policy learning, the transformer encoder translates state tokens into action tokens, which are then decoded into continuous actions using the VQVAE decoder.

\begin{algorithm}[tb]
\caption{Modeling DART for continuous action space.}
\label{alg:DART_continuous}
\begin{algorithmic}
\STATE \textbf{Representation Learning}
    \STATE $\rightarrow$ Sample trajectories $\tau$ $(s, a, s', r)$.
    \STATE $\rightarrow$ Use states in the form of an image to train an image tokenizer.
    \STATE $\rightarrow$ Use the collected actions to train an action tokenizer.
\STATE \textbf{World Model Learning}
    \STATE $\rightarrow$ Concatenate State tokens and action tokens for each time step.
    \STATE $\rightarrow$ Process each state and action token using the transformer decoder.
    \STATE $\rightarrow$ Optimize for the next state token, reward, and termination criterion.
\STATE \textbf{Policy Learning}
    \STATE $\rightarrow$ Transformer encoder modeling the policy processes the tokenized state to predict discrete action tokens.
    \STATE $\rightarrow$ VQVAE decoder maps the action token to continuous action.
    \STATE $\rightarrow$ Train the transformer encoder policy for the next $H$ steps using the imagined trajectories from the world model.
\end{algorithmic}
\end{algorithm}

\end{document}